\PassOptionsToPackage{table,usenames,dvipsnames}{xcolor}
\documentclass[conference]{IEEEtran}

\IEEEoverridecommandlockouts
\usepackage{xspace}
\usepackage{times}
\usepackage{graphicx}
\usepackage{amsmath}
\usepackage{amssymb}
\usepackage{listings}
\usepackage{xcolor}
\usepackage{footnote}

\makesavenoteenv{tabular}
\usepackage{tablefootnote} 
\usepackage{threeparttable} 
\usepackage[numbers]{natbib}
\usepackage{multicol}
\usepackage{tabularx}
\definecolor{Mycolor1}{HTML}{DAE8FC}
\definecolor{Mycolor2}{HTML}{FFE6CC}
\definecolor{Mycolor3}{HTML}{D5E8D4}
\usepackage{soul}
\usepackage{graphicx}
\usepackage{makecell}
\usepackage{float}
\usepackage{graphics}
\usepackage{etoolbox}
\usepackage{booktabs}
\usepackage{url}

\usepackage{subcaption}
\usepackage{algorithm}
\usepackage{algpseudocode}
\usepackage{makecell}
\usepackage{balance}

\usepackage{siunitx}
\usepackage{multirow}
\usepackage{tabularx}
\usepackage[export]{adjustbox}
\usepackage{overpic}
\usepackage{color, colortbl}
\usepackage{float}
\usepackage[bookmarks=true,urlcolor=black]{hyperref}

\definecolor{Gray}{gray}{0.95}  
\definecolor{CodeGreen}{rgb}{0,0.6,0}  
\definecolor{CodeBlue}{rgb}{0,0,0.8}  
\definecolor{CodeRed}{rgb}{0.8,0,0}  
\definecolor{CodeGray}{gray}{0.5}

\lstdefinestyle{mystyle}{
    backgroundcolor=\color{Gray},   
    commentstyle=\color{CodeGreen}\ttfamily\itshape,
    keywordstyle=\color{CodeBlue}\bfseries,
    numberstyle=\tiny\color{CodeGray},
    stringstyle=\color{CodeRed},
    basicstyle=\ttfamily\footnotesize,  
    breakatwhitespace=false,         
    breaklines=true,                 
    captionpos=b,                    
    keepspaces=true,                 
    numbers=left,                    
    numbersep=8pt,                  
    frame=single,                    
    framerule=0.5pt,
    rulecolor=\color{CodeGray},
    showspaces=false,                
    showstringspaces=false,
    showtabs=false,                  
    tabsize=2
}

\newcommand{\ours}{SHIFT\xspace}%{SHAFT}

\title{Uncertainty-Aware Trajectory Prediction via Rule-Regularized Heteroscedastic Deep Classification}

\author{Kumar Manas\textsuperscript{1}, Christian Schlauch\textsuperscript{2,3}, Adrian Paschke\textsuperscript{1,4},Christian Wirth\textsuperscript{2} and Nadja Klein\textsuperscript{3}
\thanks{\textsuperscript{1}Department of Mathematics and Computer Science, Freie Universität Berlin. \textsuperscript{2}Continental Automotive Technologies GmbH, AI Lab Berlin. \textsuperscript{3}Karlsruhe Institute of Technology, Scientific Computing Center, Methods for Big Data. \textsuperscript{4}Fraunhofer Institute for Open Communication Systems, Berlin, Germany. Contact: \texttt{kumar.manas@fu-berlin.de}.}
}

\begin{document}
\maketitle

\begin{abstract}
Deep learning-based trajectory prediction models have demonstrated promising capabilities in capturing complex interactions. However, their out-of-distribution generalization remains a significant challenge, particularly due to unbalanced data and a lack of enough data and diversity to ensure robustness and calibration. To address this, we propose \ours (\underline{S}pectral \underline{H}eteroscedastic \underline{I}nformed \underline{F}orecasting for \underline{T}rajectories), a novel framework that uniquely combines well-calibrated uncertainty modeling with informative priors derived through automated rule extraction. \ours reformulates trajectory prediction as a classification task and employs heteroscedastic spectral-normalized Gaussian processes to effectively disentangle epistemic and aleatoric uncertainties. We learn informative priors from training labels, which are automatically generated from natural language driving rules, such as stop rules and drivability constraints, using a retrieval-augmented generation framework powered by a large language model. Extensive evaluations over the nuScenes dataset, including challenging low-data and cross-location scenarios, demonstrate that \ours outperforms state-of-the-art methods, achieving substantial gains in uncertainty calibration and displacement metrics. In particular, our model excels in complex scenarios, such as intersections, where uncertainty is inherently higher. Project page: \url{https://kumarmanas.github.io/SHIFT/}.
\end{abstract}

\IEEEpeerreviewmaketitle
\section{Introduction}
\label{sec:intro}
\begin{figure*}[t]
    \centering
    \includegraphics[width=\textwidth]{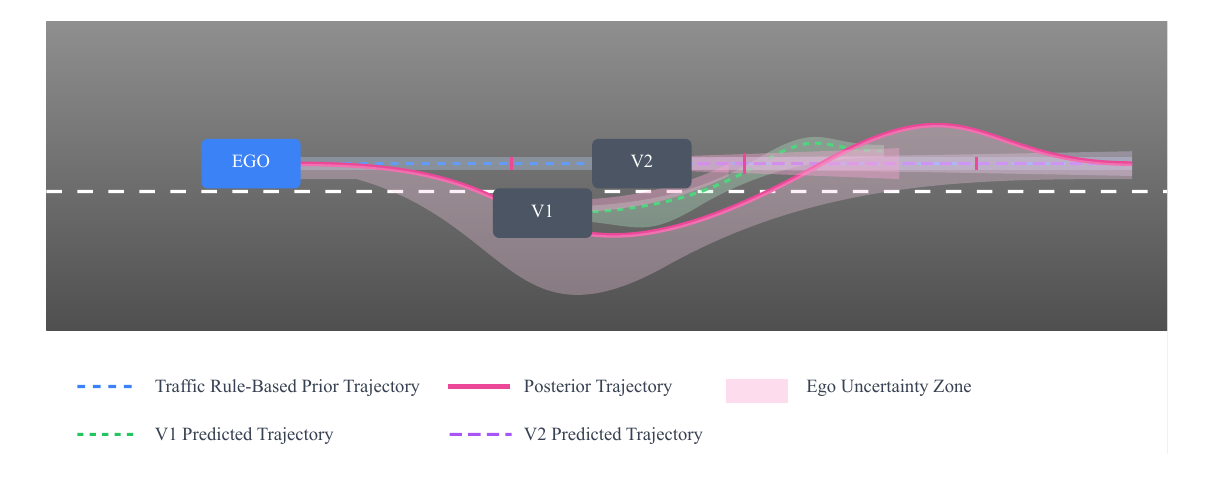}
    \caption{Visualization of \ours's trajectory prediction framework. Rule-based priors (blue dashed) generate trajectories without interaction modeling, while interaction-aware HetSNGP posteriors (pink solid) produce uncertainty-calibrated ego paths. Surrounding vehicles' predictions (V1: green dashed, V2: purple dashed) with uncertainty regions (light green/purple shading) reveal multi-agent dynamics. Denser pink stripes (not size-based) highlight where vehicle interactions amplify prediction variance, showing how our model leverages static traffic rules with real-time uncertainty propagation across all actors.}
    \label{fig:trajectory-predictions}
    \vspace{-5pt}
\end{figure*}

Trajectory prediction represents a fundamental challenge in autonomous driving and robotics, serving as a critical bridge between perception and planning systems~\citep{paden2016survey, hagedorn2023integration}. As illustrated in Fig.~\ref{fig:trajectory-predictions}, autonomous vehicles must simultaneously reason about multiple possible future trajectories while accounting for road conditions, traffic rules, and multi-agent interactions. This complex decision-making process becomes particularly challenging in urban environments, where the interactions between vehicles, pedestrians, and infrastructure are fast-changing~\citep{kretzschmar2016social, mohamed2024mpc}. Deep learning approaches for trajectory prediction have demonstrated remarkable potential in tackling these challenges~\cite{hagedorn2023integration}. Nevertheless, the safety-critical nature of autonomous driving introduces requirements that current state-of-the-art deep learning-based trajectory predictors struggle to address comprehensively. 

First, models must generalize effectively from limited data, given the high cost of data collection and long-tail distribution of unique scenarios in urban environments. For example, most observed scenarios involve relatively simple behaviors, such as following a straight road, while rare safety-critical events, like near-accidents, are particularly challenging to capture due to their low frequency. Data efficiency of deep learning-based trajectory predictors has only recently gained attention in scientific investigations~\cite{feng2024unitraj, yao2024ood}. 

Second, models must be robust to shifts in road conditions, perception systems, and geographic locations. Social norms and traffic rules provide valuable context for predicting compliant agent behavior~\cite{casas2020priorknowledge}. However, informed or knowledge-guided deep learning-based trajectory predictors have predominantly focused on physical constraints~\cite{cui2020kinematic, bahari2021injecting, li2020logic, yao2023ep}, leaving broader contextual prior knowledge underexplored~\cite{rueden2021informed}.

Third, trajectory prediction involves multiple sources of uncertainty that must be carefully quantified. Epistemic uncertainty arises from model limitations and incomplete training data, while aleatoric uncertainty stems from the stochasticity of agent behavior and environmental dynamics~\cite{huellermeier2021decomposition}. Although most trajectory prediction models address the predictive multi-modality resulting from these uncertainties (e.g., by framing it as a classification problem), many fail to provide sufficiently calibrated uncertainty estimates, leading to mode-collapse issues~\cite{bajaras2024modecollapse}.

\begin{figure*}
    \centering
    \includegraphics[width=1\linewidth]{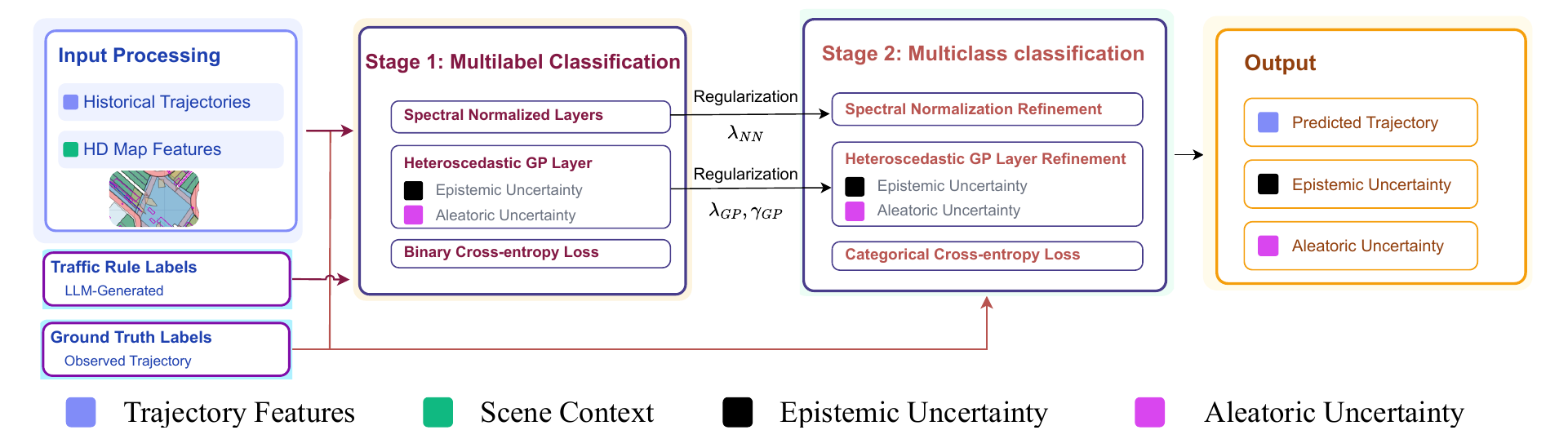}
    \caption{Concept of the \ours framework for uncertainty-aware trajectory prediction. Our model is based on the CoverNet architecture, which frames the prediction as trajectory anchor classification. The backbone is spectral-normalized, and the output layer is replaced by a heteroscedastic Gaussian process. Given the processed input, comprising trajectory histories and map features, the model is trained in a multilabel classification on traffic rule labels generated from an LLM-based rule-filtering approach (stage 1). The obtained informative prior is used to regularize the multi-class classification training on ground truth labels from the observed data (stage 2). The model outputs consist of the predicted trajectories and disentangled epistemic and aleatoric uncertainty estimates. Best viewed in color.}
    \label{fig:pipeline}
\end{figure*}

To address these requirements, we propose \ours (\underline{S}pectral \underline{H}eteroscedastic \underline{I}nformed \underline{F}orecasting for \underline{T}rajectories) to synergize uncertainty quantification with rules as soft constraints in a scalable framework as visualized in Fig.~\ref{fig:pipeline}:
\begin{enumerate}
    \item We extend the Heteroscedastic Spectral-normalized Gaussian Processes (HetSNGP) by~\citet{fortuin2021deep} to the trajectory prediction using the CoverNet baseline architecture~\cite{phan2020covernet}. CoverNet frames the prediction as a classification problem. HetSNGPs enable simultaneous estimation of epistemic and aleatoric uncertainties through a distance-aware neural GP layer combined with input-dependent noise modeling, improving calibration and requiring minimal architectural change and computational overhead.
    \item We model epistemic uncertainty while accounting for the inherent stochasticity of agent interactions through aleatoric uncertainty, enabling a sequential learning process for informative prior weight distributions. In stage 1, we encode traffic rules as synthetic training labels, capturing structured prior knowledge. These learned traffic rule priors are then imposed as regularization during stage 2, guiding model training on real-world observations. This approach soft-constrains trajectory predictions to align with traffic rules while preserving the flexibility to adapt to high-uncertainty and uncommon driving scenarios.
    \item We employ a large language model (LLM) pipeline to semi-automatically generate synthetic training labels from natural language descriptions. This approach allows us to more easily scale the integration of multiple traffic rules, either as (a) a single ``unified'' prior learned from a single knowledge task or (b) a ``chained'' prior sequentially learned from multiple knowledge tasks in our first training stage. 
\end{enumerate}

\ours increases data efficiency and robustness by integrating sets of rules into the training process. Through extensive experiments on the nuScenes dataset \citep{caesar2020nuscenes}, we demonstrate robust performance gains across varying training-set sizes (100\% vs. reduced data) and challenging geographical splits (cross-location tests). Our results show particular strength in out-of-distribution scenarios, where the combination of uncertainty awareness and rule-based soft constraints provides more reliable predictions than conventional training methods.

\section{Related Work}
\label{sec:related}

\noindent \textbf{Trajectory Prediction Models:} Existing models can be differentiated based on whether they predict the future behavior of all agents simultaneously (joint trajectory prediction) or a single agent at a time (marginal trajectory prediction), with most work focusing on the simpler latter problem \cite{hagedorn2023integration, wagner2024joint}. Deep learning approaches have achieved state-of-the-art performance in in-distribution evaluation settings, surpassing traditional methods such as kinematic models~\cite{scholler_what_2020}, Kalman filters and Bayesian Networks +~\cite{schulz2018bayesnets}.  Building on the work of \citet{hagedorn2023integration}, deep learning-based trajectory predictors can be categorized based on their input representation, model backbone, and trajectory decoding methods. Input representations include birds-eye-view (BEV) image rasterizations~\cite{phan2020covernet}, sparse vector encodings~\cite{shi2022mtr}, graphs~\cite{zhou2023qcnet}, and parametric curves~\cite{yao2023ep}. Model backbone encompass convolutional neural networks (CNN)~\cite{phan2020covernet}, recurrent neural networks~\cite{deo2020rnn}, graph neural networks~\cite{lian2020lanegcn} and more recently transformers~\cite{shi2022mtr, zhou2023qcnet, ngiam2022scenetransformer}. Trajectory decoding method include regression~\cite{ngiam2022scenetransformer}, anchor trajectory classification~\cite{phan2020covernet}, trajectory refinement~\cite{chai2019multipath} or end-point completion~\cite{gu2021densetnt}. Our chosen baseline, CoverNet~\cite{phan2020covernet}, utilizes a CNN backbone with BEV input rasterization and frames the prediction problem as anchor trajectory classification. This makes it particularly suitable for our modifications.

\noindent \textbf{Informed Trajectory Prediction:} Informed, or knowledge or rule-guided learning integrates explicit prior knowledge into the training process or architecture of deep learning models~\cite{rueden2021informed}. Most existing methods integrate physical knowledge to eliminate physically implausible outcomes, thereby simplifying the solution space~\cite{phan2020covernet, cui2020kinematic, bahari2021injecting, li2020logic, yao2023ep}. In contrast, social norms and traffic rules act as soft constraints, as they can be violated in uncertain or atypical scenarios. Their integration often relies on transfer learning~\cite{boulton2020motion} or multi-task learning methods~\cite{rahimi2024multiloss}. A related probabilistic approach learns informative priors sequentially from synthetic training labels, using them to regularize later training stages~\cite{schlauch2023ecml}. This setup offers flexibility, allowing sequences to be extended while minimizing catastrophic forgetting~\cite{delange2022continual, shwartzziv2022bayesiantransfer}. However, unlike our method, it does not incorporate traffic rules and local rule constraints. Additionally, a significant challenge remains in translating natural language descriptions of rules into synthetic training labels. Recent efforts used retrieval augmented generation (RAG)~\cite{lewis2020rag} and large language models (LLMs) to convert language descriptions into formal logic rules~\cite{CaStL_guo,manas2024cot,manas_tr2mtl}. We employ these logic rules to filter input-dependent synthetic training labels from predefined anchor trajectories.

\noindent \textbf{Uncertainty Quantification:} Mode-collapse arises when trajectory prediction models fail to adequately capture the full diversity of possible futures~\cite{bajaras2024modecollapse}. This issue is commonly addressed through design choices in trajectory decoding~\cite{hagedorn2023integration}. For instance, in anchor trajectory classification, the softmax-normalized logits represent distributional information and identify the most likely top-k anchors~\cite{phan2020covernet}. However, deterministic deep learning models often exhibit overconfidence, which hinders their ability to accurately reflect the true variance and multiple modes of the underlying distribution. Their calibration can be substantially improved using principled uncertainty estimation techniques, such as those provided by approximate Bayesian deep learning~\cite{seligmann2023bayesiandeep}. Commonly used sampling-based methods, including Deep Ensembles~\cite{seligmann2023bayesiandeep}, Monte Carlo (MC) Dropout~\cite{gal2016dropout}, Stochastic Weight Averaging Gaussian (SWAG)~\cite{maddox2019swag}, Laplace~\cite{daxberger2021laplace} or variational approximations~\cite{loo2021gvcl}, can be computationally prohibitive as they require multiple forward passes during prediction. \citet{pmlr-v205-itkina23a} demonstrated the significance of estimating epistemic uncertainty in trajectory prediction through evidential deep learning, utilizing a classification-based prediction framework similar to ours. In contrast, Spectral-normalized Gaussian Processes (SNGPs)~\cite{liu2020spectral}, belonging to the family of deterministic uncertainty estimators~\cite{charpentier2023training, postels2022detuncertainty}, provide especially compute-efficient last-layer approximation. The Heteroscedastic Spectral-normalized Gaussian Process (HetSNGP)~\cite{fortuin2021deep} leverages the heteroscedastic method for classification problems~\cite{collier2020heteroscedastic} to further improve calibration by disentangling the uncertainty into distance-aware epistemic and input-dependent aleatoric components~\cite{huellermeier2021decomposition}. Thus, unlike homoscedastic models, HetSNGP can tune the aleatoric uncertainty based on factors such as the complexity of the driving scenario at hand, instead of assuming a constant aleatoric uncertainty across all input conditions. This, in turn, can benefit the calibration of the epistemic uncertainty~\cite{collier2021calibration}, which we employ for our rule-informed learning approach.

\section{Methodology}
\label{sec:methodology}

We focus our discussion on the marginal trajectory prediction. Here, we aim to predict the future possible trajectories of a single agent at a time over some horizon, given the road topology, current state, and past trajectories of itself and its surrounding agents over some history, as well as the state of any traffic guidance systems. Our goal is to inform the model using traffic rules as explicit prior knowledge about the expected behavior of the agent.
We propose \ours to approach rule informed marginal trajectory prediction in a scalable way, as shown in Fig.~\ref{fig:pipeline}. To this end, \ours synergizes (A.) CoverNet as our baseline model for trajectory classification, (B.) HetSNGP as uncertainty-aware extension to CoverNet, that disentangles epistemic and aleatoric uncertainties, (C.) a regularization method which enables the integration of priors into our HetSNGP-CoverNet model, (D.) a sequential task setup to learn these priors from synthetic training labels, and (E.) an automated rule-filtering approach to encode these synthetic training labels from natural language descriptions. We describe these building blocks and (F.) our rule selection next.

\subsection{Trajectory Classification with CoverNet}
\label{sec:covernet}

% describe covernet architecture and anchor trajectory classification
We adopt an anchor trajectory classification approach as the foundation for our framework. This approach simplifies the trajectory prediction by generating a input-dependent or input-independent set of anchor trajectories that capture all plausible future motions~\cite{boulton2020motion,chai2019multipath}. These anchors can be represented as classes in the output space $\mathcal{Y} = \{1, 2, \dots, K\}$. The training data $\{(x_i,y_i)\}_{i=1}^N$ consists of samples $x_i$ from the $d$-dimensional input space $\mathcal{X}\in \mathbb{R}^d$ and the training labels $y_i$ being the classes in $\mathcal{Y}$ that most closely reflect the observed ground truth trajectories as measured by the Euclidean distance. The model can then be trained to classify the most likely anchor using a categorical cross-entropy loss.

Anchor trajectory classification offers several advantages. By appropriately sizing the anchor set, the prediction problem can be simplified, and the softmax-normalized logits provide probabilistic interpretations. Appropriately spaced anchors can also guard against drastic forms of mode-collapse~\cite{phan2020covernet}. However, the approach inherently introduces an irreducible approximation error, as the anchors may be arbitrarily distant from the actual ground truth trajectory. Consequently, the selection of anchor trajectories plays a critical role in the overall model performance.

For our baseline, we build on CoverNet~\cite{phan2020covernet} as it employs such an anchor classification framework and has been used in related work for informed trajectory prediction~\cite{boulton2020motion, schlauch2023ecml, schlauch2024ecai}. CoverNet processes bird’s-eye-view (BEV), agent-centric rasterized input images using a ResNet-50 backbone. This rasterization preserves spatial relationships well. Next, CoverNet generates a predefined, input-independent anchor trajectory set based on all ground truth trajectories in the training data. Depending on a single distance parameter these trajectories are clustered to an appropriate number of anchors. This set of anchors is held fixed across all inputs. Note, that CoverNet also proposes an input-dependent anchor generation based on a dynamical bicycle model which encodes physical knowledge as hard constraint. Our proposed methodology is in principle agnostic to the kind of anchor sets and can be combined with such anchor generation for physical knowledge integration. For simplicity we focus on the predefined anchor set.

\subsection{Uncertainty-Aware Trajectory Classification}
\label{sec:sngp}

\begin{figure}[ht]
    \centering
    \includegraphics[width=0.8\linewidth]{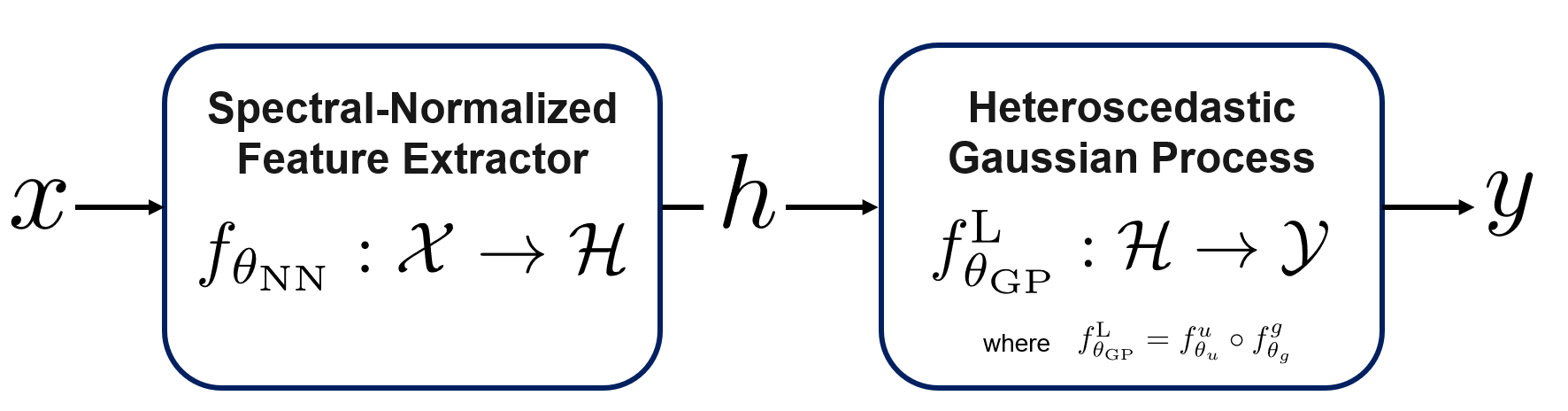}
    \caption{Components of a HetSNGP model include a spectral-normalized feature extractor and a heteroscedastic Gaussian process as output layer.}
    \label{fig:sngp}
    \vspace{-4pt}
\end{figure}

We modify CoverNet as heteroscedastic Spectral-Normalized Gaussian Process (HetSNGP)~\cite{fortuin2021deep}. The Spectral-Normalized Gaussian Process (SNGP), originally introduced by~\citet{liu2020spectral}, combines the representative power of deep neural networks with uncertainty-aware Gaussian Process (GP) models~\cite{rasmussen2006gaussian}. The architecture consists of a deterministic, spectral-normalized feature extractor $f_{\theta_{\text{NN}}}: \mathcal{X} \to \mathcal{H}$, and an hierarchical GP output layer $f^{\text{L}}_{\theta_{\text{GP}}}: \mathcal{H} \to \mathcal{Y}$  (see Figure~\ref{fig:sngp}). 

The feature extractor is a neural network parameterized through $\theta_{\text{NN}}$ that maps the high dimensional input space $\mathcal{X} \in \mathbb{R}^d $ into a low dimensional hidden space $\mathcal{H}  \in \mathbb{R}^m$ with $m \ll d$. The spectral-normalization enforces Lipschitz continuity, which can improve the generalization capability of the feature extractor \cite{liu2020spectral} and prevent feature-collapse \cite{amersfoort2021featurecollapse}. It is especially efficient to compute for the residual layers of a CNN-based backbone \cite{gouk2021lipschitz}, as used in our CoverNet baseline. The resulting training data $\{(h_i,y_i)\}_{i=1}^N$ with its embedded inputs $h_i = f_{\theta_{\text{NN}}}(x_i)$ is then fed into the GP output layer.

The GP output layer employs a radial basis function (RBF) kernel $\kappa$ and maps from the hidden space $\mathcal{H}$ into the output space $\mathcal{Y}$. Its hierarchical structure enables to quantify both, \textit{aleatoric} and \textit{epistemic} uncertainties. The GP latent variable $g$ is modeled with a zero-mean multivariate normal distribution prior per-class $c \in \mathcal{Y}$ :
\begin{equation*}
    g_{c} \sim \mathcal{N}(0, \kappa(h,h)) \quad \text{with} \quad g_{c}\in\mathbb{R}^{N \times 1}.
\end{equation*}
The posterior covariance matrix $ \kappa(h,h)$ of $g_c$  captures epistemic uncertainty. This uncertainty is class-independent, as the GP prior is shared across all classes. 

To model aleatoric uncertainty, the HetSNGP introduces a hierarchical latent variable $u$ with the following prior per-sample $i \in \{1, \dots, N\}$:
\begin{equation*}
    u_{i} \sim \mathcal{N}(g_{i}, \Sigma(h,h)) \quad \text{with} \quad u_{i}\in\mathbb{R}^{1 \times K},
\end{equation*}
where $\Sigma(h,h)$ has a low-rank approximation based on a factor covariance matrix modeled as linear neural network with parameters $\theta_u$ \cite{fortuin2021deep} to achieve scalability with the number of classes $K$. The variable $u$ captures input-dependent noise (e.g., sensor noise, agent intent ambiguity) and is processed through a heteroscedastic layer to produce aleatoric uncertainty. Unlike epistemic uncertainty, aleatoric uncertainty is sample-specific and varies across inputs. The hierarchical setup using both uncertainties correlates samples and classes in the posterior, leading to a better calibration \cite{collier2021calibration}. 

The above formulation is computationally intractable. To achieve tractability the kernel is shared between all classes and approximated as $\kappa(h,h)=\Phi \Phi^\top$ using $m$ random Fourier features (RFF) $\Phi_i = \sqrt{\frac{2}{m}}\cos(Wh_i+b)$, where $W$ and $b$ are fixed randomly sampled weights and biases respectively, based on Bochner's theorem \cite{rahimi2007bochner,liu2020spectral}. This RFF approximation reduces the computational complexity for inferring the posterior from $\mathcal{O}(N^3)$ to $\mathcal{O}(Nm^2)$ and allows us to write the GP as neural network layer with logits $g_c(h_i)=\theta_{g_c}\Phi_i$ and prior $\theta_{g_c} \sim \mathcal{N}(0, \mathbb{I})$. Since the posterior of $\theta_{g_c}$ is not of closed form, we use a Laplace approximation, which yields a Gaussian approximate posterior for the output weights $\theta_{g_c}$,
\begin{equation*}
p(\theta_{g_c}|\{(x_i,y_i)\}_{i=1}^N)  \approx \mathcal{N}(\theta_{g_c}; \theta_{g_c}^{*}, \Sigma_{g_c}), 
\end{equation*}
where $\theta_{g_c}^{*}$  is the maximum a posterior (MAP) estimate and precision $\Sigma^{-1}_{g_c}$ is given by
\begin{equation}
\label{eq:precision}
\Sigma^{-1}_{g_c} \approx \mathbb{I} + \sum_{i=1}^N p_{i,c}(1 - p_{i,c})\Phi_i\Phi_i^{\top},
\end{equation}
where $p_{i,c}$ is the softmax output $p(y_i = c | u^{*}_{i,c})$ given the per-class per-sample $u^{*}_{i,c}=\theta_{g_c}^{*} \Phi^\top $. Overall, the trainable weights in our output layer consist of $\theta_\text{GP}=\{\theta_u, \theta_{g_c} | \forall c \in \mathcal{Y}\}$. We adopt this output layer for our CoverNet baseline by replacing its dense output layer.

During inference, we approximate the marginal predictive distribution $p(y|h) = \int p(y|u)p(u|h)du$ of \( u \) using Monte Carlo (MC) sampling:
\begin{equation*}
\begin{split}
p(y\mid h) &= \mathbb{E}_{p(\theta_{\text{GP}} \mid \{(h_i, y_i)\}_{i=1}^N)} \big[p(y \mid h, \theta_{\text{GP}})\big] \\
&\approx \frac{1}{S} \sum_{s=1}^{S} p(y \mid h, \theta_{\text{GP}}^{(s)}),
\end{split}
\end{equation*}
where \( S \) is the number of MC samples, and \( \theta_{\text{GP}}^{(s)} \) are samples drawn from the approximate posterior distribution. Following Fortuin et al. (2021), a temperature parameter \( \tau \) is introduced to recalibrate uncertainty at test time. The MC sampling is computationally efficient, as it involves sampling only from the output layer and supports parallelization. %, and provides robust uncertainty estimates over the \( K \) anchor trajectories. 

\subsection{Regularization using Informative  
Priors}
\label{sec:regularization}

To enable the integration of traffic rule as informative priors, we leverage the regularization method by~\citet{schlauch2024ecai}. The idea can be seen as an extension of online elastic weight consolidation~\cite{online_EWC} for GPs with RFF-approximated RBF-kernels. Assume we have some informative prior for the GP output layer $\pi = \mathcal{N}(\theta_{g_c}; \tilde{\theta}_{g_c}, \tilde{\Sigma}_{g_c})$ rather than a standard Gaussian prior as before. We then regularize the MAP estimate $\theta_{g_c}^{*}$ through this prior. This leads to the penalized log-likelihood
\begin{equation}
        -\log{p_{\theta_{g_c}}(y_{i} | h_{i})} - \frac{\lambda_\text{GP}}{2}(\theta_{g_c} - \tilde{\theta}_{g_c})^\top \tilde{\Sigma}_{g_c}^{-1}(\theta_{g_c} - \tilde{\theta}_{g_c}). 
\end{equation}
The posterior variance of $\theta_{g_c}$ is then given by \begin{equation}
   \Sigma_{g_c}^{-1} \approx \gamma_\text{GP} \tilde{\Sigma}_{g_c}^{-1} + \sum_{i=1}^N p_{i,c}(1 - p_{i,c})\Phi_i\Phi_i^{\top}.
\end{equation}
This regularization scheme introduces two new hyperparameters $\lambda_\text{GP}$ and $\gamma_\text{GP}$, which we assume are the same for all classes. The hyperparameter $\lambda_\text{GP}>0$ tempers the overall prior, while the hyperparameter $0<\gamma_\text{GP}<1$ controls the learning decay when the model is sequentially trained \cite{online_EWC}. 

In addition, assume we have some informative prior for the feature extractor $\pi=\mathcal{N}(\theta_{\text{NN}}; \tilde{\theta}_{\text{NN}}, \mathbb{I})$ too. Regularizing the feature extractor is then equivalent to $\mathcal{L}2$-regularization for the MAP estimates $\theta^{*}_\text{NN}$, obtained by minimizing
\begin{equation}
        -\log{p_{\theta_\text{NN}}(y_{i} | h_{i})} - \frac{\lambda_\text{NN}}{2}(\theta_\text{NN} - \tilde{\theta}_{\text{NN}})^2,
\end{equation}
which introduces an additional hyperparameter $\lambda_\text{NN}$ controlling the strength of the parameter binding.
Note, that we do not specifically place an informative prior on the parameters $\theta_u$ of the GP output layer as this head captures a property that is purely related to the data itself.

\subsection{Learning Informative Traffic Rule Priors %Informative Priors
}
\label{sec:rule}

We encode prior knowledge about traffic rules by incorporating additional tasks with synthetic training labels~\cite{schlauch2023ecml}. Given the input samples $x_i \in \mathcal{X}$ and the set of anchor trajectories, we define the synthetic training labels $y_i$ as those classes in $\mathcal{Y}$ that yield rule-compliant anchor trajectories. In this multi-label classification setup, the model is trained to predict which anchors satisfy the respective traffic rule using a binary cross-entropy loss. 

\begin{figure}[htbp]
    \centering
    \includegraphics[width=0.8\linewidth]{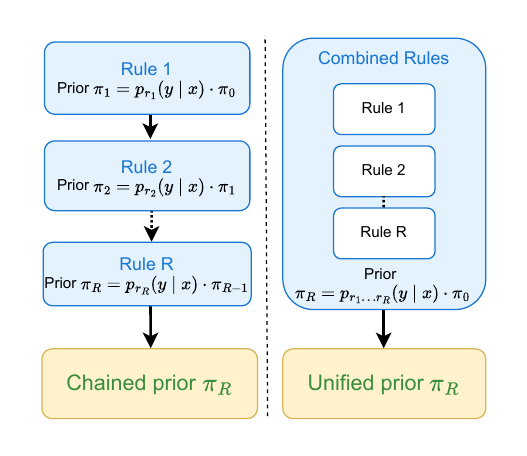}
    \caption{Illustration of the possible task setups in the first training stage. The uninformed prior $\pi_{0}$ is either sequentially updated using distinct traffic rule labels $p_{r_l}$ for each rule $l=1,\ldots,R$ or updated once from a unified traffic rule $p_{r_1,\ldots,r_R}$ simultaneously. We denote the obtained final prior ${\pi_{R}}$ as chained and unified priors, respectively. Both setups can be mixed. The obtained final prior is used to regularize the training on the observed ground truth trajectories in the next training stage.}
    \label{fig:chained}
\end{figure}

Using an uncertainty-aware model in \ours, we can sequentially train on multiple tasks. Starting with an uninformative prior $\pi_0$,  the posterior distribution from a previous task serves as informative prior $\pi_{l}$ to regularize the subsequent task. The probabilistic regularization as described in Sec.~\ref{sec:regularization} constitutes a key advantage over conventional transfer learning, which does not guard against catastrophic forgetting~\cite{delange2022continual} or overly biasing subsequent training tasks \cite{shwartzziv2022bayesiantransfer}. Another important advantage lies in the flexible task setup, as visualized in Fig.~\ref{fig:chained}. Thereby, we can define a single task that encodes prior knowledge about multiple traffic rules $p_{r_1,\ldots,r_R}(y|x)$ at the same time, learning a ``unified prior'' (right). Alternatively, we can define multiple tasks that encode prior knowledge about one traffic rule $p_{r_l}(y|x)$ each, learning a ``chained prior'' (left).  This sequential task setup allows us to re-use priors easily, constituting an advantage over multi-task joint learning setups \cite{rahimi2024multiloss}.

\subsection{Synthetic Training Labels from Traffic Rules}
\label{automated_prior}
\begin{figure}
    \centering
    \includegraphics[width=0.9\linewidth]{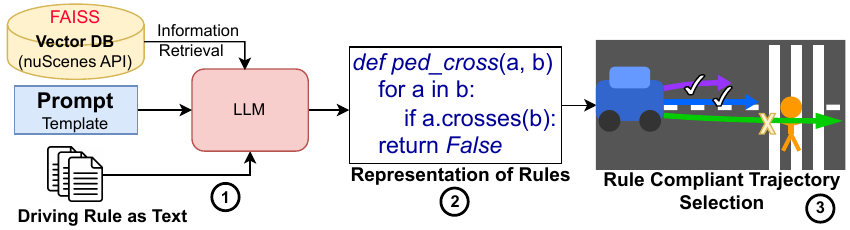}
    \caption{High-Level LLM pipeline for synthetic training label creation:(1) rules, prompt, and nuScenes API are LLM inputs, (2) the LLM generates Python functions of rules and human experts review generated functions by running test cases using a sample traffic scene, (3) based on rules synthetic labels for training are generated.}
    \label{fig:LLM_overview}
    \vspace{-10pt}
\end{figure}

To scale the integration of traffic rules, we employ a semi-automated filtering approach that (a) translates natural language descriptions into executable Python functions using a LLM-based human-in-the-loop setup and (b) uses these functions to label each trajectory in an anchor set for rule compliance within a given scene. The first step (a) utilizes the retrieval-augmented generation (RAG) technique for the few-shot prompting of the LLM. We extended methodology developed by~\citet{manas2024cot} using Llama3.3\footnote{\href{https://ollama.com/library/llama3.3}{https://ollama.com/library/llama3.3}} model for this use case. When provided with a natural language rule description, RAG retrieves semantic elements from a dataset API. The LLM then integrates the relevant API calls to generate an executable Python function grounded in dataset API, which the user evaluates. If the function meets expert expectations, it is used in the second step (b) to label the rule compliance of the trajectory anchor set for all samples in the training dataset. Fig.~\ref{fig:LLM_overview} illustrates this process for synthetic label creation.

This pipeline accelerates rule integration, reducing the need for extensive manual coding or rule formalization. Various natural language traffic rules can be translated depending on the dataset API's capabilities and the available semantic information (e.g., lane types). However, even with sufficient semantic information, evaluating crucial temporal aspects (e.g., safety distances based on the velocities of multiple agents over time) remains challenging for the LLM. Further details on the LLM prompts and limitations are provided in Appendix~\ref{appendix:llm_and_prompt}.

\subsection{Rule Selection}
\label{sec:rules}

We select two sets of rules to allow comparisons to previous work and demonstrate the scalability of \ours. These sets of rules model both globally-applicable and situational behavioral constraints, while avoiding the limitations of our LLM-powered rule-filtering approach. By integrating rules of varying complexity, we can account for more nuanced interactions with the traffic infrastructure.

Our first set contains a globally-applicable traffic rule: \begin{itemize} \item \textit{Stay within the road boundaries}: This rule enforces the fundamental constraint of staying within drivable areas, reflecting high-level driving behavior. 
\end{itemize}

Our second set of rules are situational stop-related traffic rules, including: \begin{itemize} \item \textit{Stopping at red signals}: A rule that prohibits crossing traffic light zones when the red light is active, ensuring compliance with traffic signals. 
 \item \textit{Respecting right-of-way}: A rule that mandates yielding to other participants at yield or stop signs, capturing critical right-of-way interactions.
 \item \textit{Prioritizing pedestrian safety}: A rule that requires giving way to pedestrians at active crossings, emphasizing safety in shared spaces.   
\end{itemize}

\section{Experimental Design}
\label{sec:exp_results}
\setlength{\tabcolsep}{6pt} % Adjust column spacing
\renewcommand{\arraystretch}{1.25}
\begin{table*}[!tbp]
\caption{Comparison of \ours and baselines on Full Dataset of nuScenes. Our approach is compared against other uncertainty-aware classification baselines. Bold indicates best, and second best is underlined. Lower is better for metrics.
}
\label{tab:full_data}
\centering

\begin{tabular}{lcccccc}
\hline
\textbf{Model} & \textbf{minADE$_{1}$ $\downarrow$} & \textbf{minADE$_{5}$ $\downarrow$} & \textbf{minFDE$_{1}$ $\downarrow$} & \textbf{NLL $\downarrow$} & \textbf{ECE $\downarrow$} & \textbf{RNK $\downarrow$} \\
\hline
CoverNet\footnotemark~\cite{phan2020covernet}   & 4.92 $\pm$ \tiny 0.15      & 2.34 $\pm$\tiny 0.05  & 10.94 $\pm$ \tiny 0.27  & 3.47 $\pm$ \tiny 0.06   & 0.05 $\pm$ \tiny 0.02   & 15.55 $\pm$ \tiny 0.73   \\
GVCL-Det~\cite{schlauch2023ecml}         & 4.55 $\pm$ \tiny 0.11 & 2.26 $\pm$ \tiny 0.05 & 9.93 $\pm$ \tiny 0.39 & 3.60 $\pm$ \tiny 0.08 & 0.18 $\pm$ \tiny 0.02 & 14.88 $\pm$ \tiny 0.94  \\
SNGP$_U$ (Without Rules)~\cite{schlauch2024ecai}  & 4.53 $\pm$ \tiny 0.09 & 2.25 $\pm$ \tiny 0.04 & 10.31 $\pm$ \tiny 0.27 & 3.23 $\pm$ \tiny 0.01 &\underline{0.03} $\pm$ \tiny 0.01 & 13.25 $\pm$ \tiny 0.19   \\
SNGP (With Rules)     & 4.31 $\pm$ \tiny 0.02       & 2.17 $\pm$ \tiny 0.01       & 9.78 $\pm$ \tiny 0.05      & \textbf{3.15} $\pm$ \tiny 0.01   & \underline{0.03} $\pm$ \tiny 0.01   & 11.93 $\pm$ \tiny 0.11  \\
\hline
SHIFT (Ours With Chained Prior)     & \textbf{4.11} $\pm$ \tiny 0.05 & \textbf{2.12} $\pm$ \tiny 0.01 & \underline{9.24} $\pm$ \tiny 0.12 & \underline{3.16} $\pm$ \tiny 0.02 & \textbf{0.02} $\pm$ \tiny 0.01 & \underline{11.85} $\pm$ \tiny 0.22 \\
SHIFT (Ours With Unified Prior)     & \underline{4.15} $\pm$ \tiny 0.04 & \underline{2.13} $\pm$ \tiny 0.05 & \textbf{9.23} $\pm$ \tiny 0.11 & \textbf{3.15} $\pm$ \tiny 0.03 & \textbf{0.02} $\pm$ \tiny 0.01 & \textbf{11.52} $\pm$ \tiny 0.45 \\
\hline
\end{tabular}
\vspace{-13pt}
\end{table*}

We evaluate the model's performance in various scenarios, including standard full dataset training, low data regimes, and out-of-distribution generalization across different geographical locations within the nuScenes~\cite{caesar2020nuscenes} dataset.

\subsection{Dataset} We conduct our experiments on the nuScenes dataset \citep{caesar2020nuscenes}. We selected it for geographic diversity (Boston, USA, and Singapore), semantic information (including drivable areas, stop lines, walkways), and richly annotated agent trajectories. Its geographic split enables controlled out-of-distribution (OoD) evaluation, where models trained on data from one region (e.g., Boston’s grid-like roads with right-side driving) are tested on another (e.g., Singapore’s curved intersections and left-side driving). This mimics real-world deployment challenges where models encounter unseen environmental semantics. For our full dataset experiments, we utilize the train/val/test splits following \citet{phan2020covernet}. In terms of location diversity, roughly $39\%$ of the scenes are set in Singapore, while $61\%$ take place in Boston. Refer to Appendix~\ref{appendix:dataset} for more details about the dataset.

\subsection{Evaluation Metrics} We evaluate our approach using three categories of metrics: (1) Displacement Metrics, measuring spatial accuracy of predictions; (2) Distribution-aware Metrics, quantifying the quality of predicted probability distributions; and (3) Ranking Metrics, assessing the model's ability to assign higher probabilities to more likely trajectories.
\textbf{minADE$_k$} (Minimum Average Displacement Error): Measures the average $l_2$ distance in meters between the ground truth trajectory and the closest predicted trajectory among the top-k predictions. We report both minADE$_1$ and minADE$_5$ to evaluate single-mode and multi-mode prediction accuracy respectively. \textbf{minFDE} (Minimum Final Displacement Error): measures the minimum $l_2$ distance between the predicted final position and the ground truth final position among the top-k predictions. \textbf{Negative Log-Likelihood (NLL)}: Quantifies the quality of the predicted probability distribution by measuring the negative log-likelihood of the ground truth trajectory under the model's predictions. A lower NLL indicates better alignment between the predicted distribution and observed trajectories.
\textbf{Expected Calibration Error (ECE)}: Measures the alignment between predicted probabilities and empirical frequencies~\cite{ECE_metric_2015}. ECE computes the weighted average of the absolute difference between predicted probabilities and observed frequencies across predefined probability bins. A lower ECE indicates better calibration, where the model's predicted probabilities reliably correspond to observed frequencies. \textbf{Rank (RNK)}: Measures the rank of the ground truth trajectory among all predicted trajectories when sorted by their predicted probabilities~\cite{rnk_metrics}. This metric also measures the calibration of the anchor classification, which can be equally understood as a ranking problem. A lower rank indicates the model assigns higher probabilities to trajectories closer to the ground truth.

%data subsample result
\renewcommand{\arraystretch}{1.3}
\begin{table*}[!ht]
\centering
\footnotesize
\setlength{\tabcolsep}{3pt}
\caption{Impact of Reduced Training Data on Performance Metrics. Bold values indicate the best results.}
\label{tab:reduced_data_combined}
\begin{tabular}{l|cc|cc|cc|cc|cc}
\hline
\textbf{Model} & \multicolumn{2}{c|}{\textbf{minADE$_{1}$} $\downarrow$} & \multicolumn{2}{c|}{\textbf{minADE$_{5}$} $\downarrow$} & \multicolumn{2}{c|}{\textbf{minFDE$_{1}$} $\downarrow$} & \multicolumn{2}{c|}{\textbf{NLL} $\downarrow$}  & \multicolumn{2}{c}{\textbf{RNK} $\downarrow$} \\[-0.1ex]
Train Data Used(in \%) $\rightarrow$ & 50\% & 10\% & 50\% & 10\% & 50\% & 10\% & 50\% & 10\% & 50\% & 10\% \\
\hline
SNGP$_U$ (Without Rules) & 4.57$\pm$\tiny 0.05 & 5.00$\pm$\tiny 0.04 & 2.26$\pm$\tiny 0.04 & 2.52$\pm$\tiny 0.03 & 10.40$\pm$\tiny 0.15 & 11.36$\pm$\tiny 0.22 & 3.30$\pm$\tiny 0.02 & 3.60$\pm$\tiny 0.02 & 14.62$\pm$\tiny 0.17 & 25.19$\pm$\tiny 0.30 \\

SNGP   (With Rules) & 4.41$\pm$\tiny 0.04 & 4.69$\pm$\tiny 0.07 & 2.21$\pm$\tiny 0.03 & 2.32$\pm$\tiny 0.05 & 9.95$\pm$\tiny 0.09 & 10.62$\pm$\tiny 0.16 & 3.23$\pm$\tiny 0.02 & 3.47$\pm$\tiny 0.02 & 13.10$\pm$\tiny 0.11 & 17.80$\pm$\tiny 0.41 \\
SHIFT (Ours Unified Prior) & \textbf{4.28}$\pm$\tiny 0.05 & \textbf{4.60}$\pm$\tiny 0.06 & \textbf{2.15}$\pm$\tiny 0.04 & \textbf{2.30}$\pm$\tiny 0.03 & \textbf{9.60}$\pm$\tiny 0.15 & \textbf{10.21}$\pm$\tiny 0.14 & \textbf{3.18}$\pm$\tiny 0.04 & \textbf{3.45}$\pm$\tiny 0.03 & \textbf{12.19}$\pm$\tiny 0.5 & \textbf{17.20}$\pm$\tiny 0.53 \\
\hline
\end{tabular}
\vspace{-5pt}
\end{table*}

\subsection{Baselines and Implementation}

We compare \ours against CoverNet-based classification baselines from previous work, as these best illustrate the impact of the rule integration and last layer modifications. Therefore, our selected baselines include the conventional CoverNet~\citep{phan2020covernet}, Deterministic Generalized Variational Continual Learning (GVCL-Det)~\citep{schlauch2023ecml} with CoverNet, an uninformed SNGP-CoverNet model (SNGP$_U$)~\citep{schlauch2024ecai}, and a rule-informed SNGP-CoverNet model (SNGP with Rules) that integrates sets of rules similar to the unified prior setup of \ours. The SNGP baselines are uncertainty-aware but do not disentangle uncertainty components, serving as The SNGP baselines are uncertainty-aware but do not disentangle uncertainty components, serving as an ablation to our model. The SNGP model only uses epistemic uncertainty.

Regarding the implementation, we use a ResNet50 backbone and a predefined anchor set of $415$ modes for all evaluated models. The input is a rasterized BEV (480x480 pixel) image and color-coded drivability area, lanes, walkways, stop lines and agent trajectories as described by \citet{phan2020covernet}. We encode the agent trajectories with \textit{1-second} observation window and \textit{6-second} prediction horizon. This design aligns with evidence that short-term observations dominate trajectory dynamics, contributing to 80\% of the predictive impact as noted in~\citet{scholler_what_2020}. This mitigate tracking error propagation and improve OoD generalization through reduced temporal dependencies~\cite{monti_how_2022,becker_red_2019}.

\begin{table*}[!tbp]
\centering
\caption{\small Comparison of \ours for Geographic Generalization. The best results are highlighted in bold. Both in-distribution (ID) and out-of-distribution (OoD) performance are evaluated to assess the model's ability to generalize to new driving scenarios and to quantify performance degradation when encountering unseen environments.}
\label{tab:geo_generalization}
\begin{tabular}{llcccccc}
\hline
\textbf{Region Pair (Trained on $\to$ Tested on)} & \textbf{Model} & \textbf{minADE$_{1}$ $\downarrow$} & \textbf{minADE$_{5}$ $\downarrow$} & \textbf{minFDE$_{1}$ $\downarrow$} & \textbf{NLL $\downarrow$} & \textbf{RNK $\downarrow$} \\
\hline
Boston $\to$ Boston (ID)
                    & SNGP$_U$ (Without Rules) & 4.50{\tiny $\pm$ 0.04}       & 2.19{\tiny $\pm$ 0.02}       & 10.13{\tiny $\pm$ 0.11}       & 3.31{\tiny $\pm$ 0.06}   & 13.64{\tiny $\pm$ 1.31}   \\
                    & SNGP (With Rules)     & 4.39{\tiny $\pm$ 0.05}       & 2.15{\tiny $\pm$ 0.03}       & 9.92{\tiny $\pm$ 0.11}       & 3.22{\tiny $\pm$ 0.02}   & 12.56{\tiny $\pm$ 0.14}   \\
                    & SHIFT (Ours With Unified Prior)     & \textbf{4.27{\tiny $\pm$ 0.02}} & \textbf{2.14{\tiny $\pm$ 0.03}} & \textbf{9.61{\tiny $\pm$ 0.04}} & \textbf{3.19{\tiny $\pm$ 0.01}} & \textbf{12.23{\tiny $\pm$ 0.22}} \\
\hline
Singapore $\to$ Singapore (ID) 
                       & SNGP$_U$ (Without Rules) & 4.43{\tiny $\pm$ 0.07}      & 2.20{\tiny $\pm$ 0.06}       & 9.83{\tiny $\pm$ 0.15}       & 3.31{\tiny $\pm$ 0.06}   & 13.64{\tiny $\pm$ 1.31}   \\
                       & SNGP (With Rules)    & 4.35{\tiny $\pm$ 0.04}      & 2.19{\tiny $\pm$ 0.02}       & 9.71{\tiny $\pm$ 0.14}       & 3.29{\tiny $\pm$ 0.01}   & 12.95{\tiny $\pm$ 0.27}   \\
                       & SHIFT (Ours With Unified Prior)    & \textbf{4.28{\tiny $\pm$ 0.03}} & \textbf{2.16{\tiny $\pm$ 0.06}} & \textbf{9.45{\tiny $\pm$ 0.08}} & \textbf{3.23{\tiny $\pm$ 0.04}} & \textbf{12.46{\tiny $\pm$ 0.51}} \\
\hline
Boston $\to$ Singapore (OoD)
                       & SNGP$_U$ (Without Rules) & 4.82{\tiny $\pm$ 0.07}      & 2.60{\tiny $\pm$ 0.07}       & 10.95{\tiny $\pm$ 0.18}       & 3.52{\tiny $\pm$ 0.04}   & 18.39{\tiny $\pm$ 0.66}   \\
                       & SNGP (With Rules)    & 4.65{\tiny $\pm$ 0.07}      & 2.48{\tiny $\pm$ 0.09}       & 10.57{\tiny $\pm$ 0.21}       & 
                       \textbf{3.46{\tiny $\pm$ 0.04}}& 17.95{\tiny $\pm$ 0.58}   \\
                       & SHIFT (Ours With Unified Prior)    & \textbf{4.48{\tiny $\pm$ 0.05}} & \textbf{2.41{\tiny $\pm$ 0.05}} & \textbf{10.09{\tiny $\pm$ 0.02}} & 3.47{\tiny $\pm$ 0.04} & \textbf{16.62{\tiny $\pm$ 1.1}}  \\
\hline
Singapore $\to$ Boston (OoD)
                       & SNGP$_U$ (Without Rules) & 5.36{\tiny $\pm$ 0.07}      & 2.68{\tiny $\pm$ 0.08}       & 12.18{\tiny $\pm$ 0.26}       & 3.65{\tiny $\pm$ 0.01}   & 21.56{\tiny $\pm$ 1.40}   \\
                       & SNGP (With Rules)    & 5.17{\tiny $\pm$ 0.06}      & 2.63{\tiny $\pm$ 0.08}       & 11.76{\tiny $\pm$ 0.25}       & 3.60{\tiny $\pm$ 0.05}   & 19.34{\tiny $\pm$ 1.03}   \\
                       & SHIFT (Ours With Unified Prior)    & \textbf{4.97{\tiny $\pm$ 0.14}} & \textbf{2.50{\tiny $\pm$ 0.05}} & \textbf{11.21{\tiny $\pm$ 0.36}} & \textbf{3.54{\tiny $\pm$ 0.04}} & \textbf{17.55{\tiny $\pm$ 0.48}} \\
\hline
\end{tabular}
\end{table*}
\subsection{ Experiments}
\label{subsec:eval_protocol}
To ensure a reliable comparison with the baselines, each experiment is conducted over five independent runs and we report the mean and standard deviation of the results. We evaluate \ours with a unified prior integrating both sets of traffic rules in the following experiments:
\begin{itemize}
    \item \textbf{Full Data}: We compare \ours against all baselines using the full nuScenes train-val-test splits to evaluate general performance. We also compare against a version with a chained prior, integrating all traffic rules sequentially, to investigate the flexibility of the task setup in our first training stage.
    \item \textbf{Reduced Data}: We compare \ours against the SNGP baselines on training data subsets, using 50\% and 10\% of training samples respectively, to assess the data efficiency.
    \item \textbf{Geographic Generalization}:
    We compare \ours with a unified prior against the SNGP baselines on location shifts to assess OoD generalization. These experiments include an \textit{in-distribution (ID)} training/testing in the same location as baseline (\texttt{Boston} $\to$ \texttt{Boston} and \texttt{Singapore} $\to$ \texttt{Singapore}) and an \textit{out-of-distribution (OoD)} training/testing across locations ( \texttt{Boston} $\to$ \texttt{Singapore} and \texttt{Singapore} $\to$ \texttt{Boston}).
    \item \textbf{Rule Ablation}:
    We compare \ours against variants that incorporate only one of the rule sets, as well as a version without any rules.
\end{itemize}
 
\footnotetext{We use the result reported in~\citet{schlauch2024ecai}, where they implemented CoverNet with \SI{1}{\second} of history observation and \SI{6}{\second} prediction horizon along with uncertainty integration.}
These experiments test data efficiency and robustness to geographic distribution shifts and assess uncertainty calibration within these contexts. We also report qualitative visualization and prediction latency. In Appendix~\ref{appendix:result_additional} we highlight results related to the impact of the posterior GP.
%This protocol tests robustness to \textbf{low-data regimes}, \textbf{geographic distribution shifts}, and \textbf{uncertainty calibration} under domain gaps. 
\section{Results and Discussions}
\label{exp:results}

\subsection{Full Dataset}
Table~\ref{tab:full_data} presents the performance comparison on the full dataset. Among the \ours configurations, the version with a unified prior achieves the best minFDE$_1$ and RNK scores, while the chained prior setup excels in minADE$_1$ and minADE$_5$. More importantly, both variants of \ours significantly outperform baseline methods across key metrics,  especially in distribution-aware metrics such as the ECE. These results highlight two properties of \ours. First, the task setup in the first training stage is flexible enough to accommodate both chained and unified priors without substantial performance trade-offs. Second, the improved calibration of the heteroscedastic GP layer directly impacts the effectiveness of the regularization-based approach.

\subsection{Reduced Datasets}
Table~\ref{tab:reduced_data_combined} presents the model performance with reduced training data (50\% and 10\% of the original dataset). Even in data-scarce environments, \ours consistently outperforms baseline methods across all metrics, demonstrating its robustness in low-data regimes. These results highlight the importance of incorporating traffic rules as prior knowledge in autonomous driving, where data collection is costly, and models must generalize effectively.

\subsection{Geographic Generalization}
Geographic generalization results in Table~\ref{tab:geo_generalization} further validate the effectiveness of \ours. In in-distribution (ID) scenarios (Boston $\to$ Boston and Singapore $\to$ Singapore), our model outperforms the baselines, demonstrating improved prediction accuracy (minADE$_1$, minADE$_5$ and minFDE$_1$) and better-calibrated uncertainty quantification (NLL and RNK). In out-of-distribution (OoD) scenarios (Boston $\to$ Singapore and Singapore $\to$ Boston), \ours demonstrates enhanced robustness, significantly reducing minADE$_1$, minFDE$_1$, and RNK compared to the baselines. The Singapore $\to$ Boston transfer exhibits a more pronounced improvement over the baseline, likely due to Singapore’s challenging and diverse data providing a stronger generalization capacity. In contrast, the Boston $\to$ Singapore transfer shows a more modest gain over SNGP (with Rules), possibly due to limited exposure to Singapore like complex traffic patterns. Nevertheless, the results demonstrate that \ours mitigates performance degradation under OoD conditions.
\subsection{Rule Ablation}
\label{sec:ruleablation}

We analyze the role of traffic rules in the \ours model by isolating two rule sets—drivability rules and stop-related rules—as outlined in Sec.~\ref{sec:rules}. Since stop-related rules (e.g., red light, right-of-way, and pedestrian priority) apply only to a subset of test cases, we group them together in this ablation study to assess their combined impact. The results, shown in Table~\ref{table:ablation_study}, indicate that stop-related rules play a more significant role in improving top-k displacement metrics, achieving the lowest minADE$_1$ (4.10) and minADE$_5$ (2.13). However, drivability rules, despite having a limited effect, improve calibration by lowering NLL from 3.19 (no rules) to 3.20. This suggests that while drivability constraints may not strongly influence top-k accuracy, they help refine predictive uncertainty. ``Only Drivability Rule” refers to global road boundary rules, while ``Without Traffic Rules” means neither global nor local
rules were used.

Importantly, the unified prior (combining both rule sets) achieves the best overall performance, with improvements in both accuracy (minADE$_5$: 2.13, minFDE$_1$: 9.23) and calibration (NLL: 3.15, RNK: 11.52). This highlights the complementary nature of global constraints (drivability) and situational constraints (stop-related rules) in learning informative priors across all test cases, especially in distribution-aware metrics. Comparing SNGP with \ours, we observe that even without rule priors, our model significantly outperforms both the rule-based and rule-free SNGP baselines.

\begin{table}[ht!]
\centering
\caption{Ablation Study: Traffic Rules Impact on \ours.}
\vspace{-5pt}
\resizebox{\columnwidth}{!}{%
\begin{tabular}{@{}l|ccccc@{}}
\toprule
\textbf{Configuration} & \textbf{minADE$_1$} $\downarrow$ & \textbf{minADE$_5$} $\downarrow$ & \textbf{minFDE$_1$} $\downarrow$ & \textbf{NLL} $\downarrow$  & \textbf{RNK} $\downarrow$ \\
\midrule
SHIFT (Unified Prior) & 4.15 & \textbf{2.13} & \textbf{9.23} & \textbf{3.15} &\textbf{11.52} \\
SHIFT (Only Stop Rules) & \textbf{4.10} & \textbf{2.13} & 9.28 & 3.17 &12.10 \\
SHIFT (Only Drivability Rule) & 4.25 & 2.22 & 9.63 & 3.20 & 12.34\\
SHIFT (Without Traffic Rules) & 4.22 & 2.16 & 9.57 & 3.19 & 12.73\\
SNGP$_U$ (Without Rules) & 4.53 & 2.25 & 10.31 & 3.23 & 13.25\\

\bottomrule
\end{tabular}
}
\vspace{-15pt}
\label{table:ablation_study}
\end{table}
\begin{figure*}[h]
    \centering
    \includegraphics[width=1\linewidth]{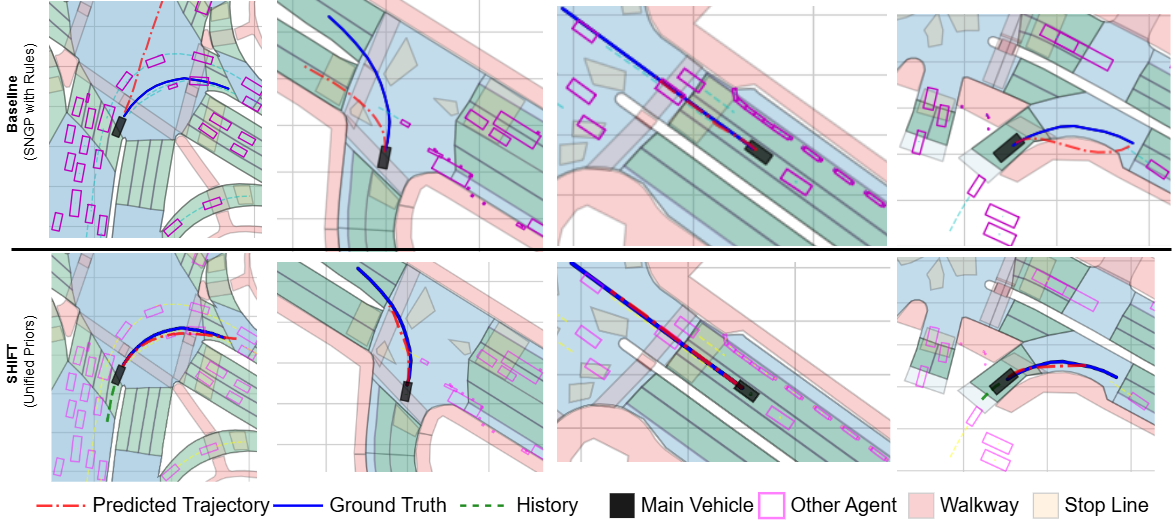}
    \caption{\textbf{Qualitative Comparison of Trajectory Predictions.} The first row displays predictions from our best baseline model, while the second row presents results from \ours. Each column represents a different scene. The target agent, for which predictions are made, is highlighted with a black rectangle. The past trajectory (shown only for \ours) is also included for reference. Best viewed in color.}
    \label{fig:comparison}
    \vspace{-10pt}
\end{figure*}
\subsection{Qualitative Results}
\label{ssub:qualitative_result}
In Fig.~\ref{fig:comparison} we visually compare predictions of \ours and SNGP with rules. At intersections, where agent behavior is inherently stochastic, \ours more accurately predicts the correct turn direction. Similarly, in scenarios involving stopping areas and multiple nearby vehicles, our model's predictions align more closely with the ground truth trajectory than those of the baseline. Additional visualizations are available in the Appendix~\ref{appendix:qualitative_results}. Notably, in overtaking scenarios, both models performed comparably.

\subsection{Runtime Consideration}
\label{runtime}
Computational efficiency is crucial for autonomous driving applications. Our benchmarks indicate that \ours achieves an average prediction latency of \SI{7.4}{\milli\second} per sample on a single NVIDIA RTX A5000, compared to \SI{5.6}{\milli\second} for CoverNet. This demonstrates that \ours incurs only a minimal computational overhead while enhancing trajectory prediction capabilities.
\section{Conclusion} 
\label{sec:conclusion}

We have introduced \ours, an uncertainty-aware trajectory predictor that integrates prior knowledge of traffic rules into deep learning models through a scalable probabilistic formulation. By explicitly modeling disentangled heteroscedastic uncertainty, \ours captures both aleatoric and epistemic uncertainties, enabling a more accurate representation of agent behavior while reducing overconfidence. Our empirical evaluations on nuScenes demonstrate that \ours produces not only more accurate predictions but also better-calibrated uncertainty estimates, due to the incorporation of rules as both chained and unified priors. The demonstrated flexibility allows end users to incrementally train and adapt the model by integrating new, user-defined driving rules as needed. Such adaptability makes \ours particularly well-suited for safety-critical applications in autonomous driving, where agent behaviors are governed by complex social norms and traffic regulations. In future work, we aim to extend this framework to incorporate interactive priors for multi-agent coordination and validate its efficacy within closed-loop planning pipelines.
\section{Limitations}  
\label{sec:limitations}  

\ours demonstrates promising results by integrating traffic rules into trajectory prediction. However, several open research questions remain. 
First, \ours so far been demonstrated only in a trajectory classification context, where compliance is evaluated based on predefined trajectory anchors. Adapting it to other trajectory decoding strategies, such as regression, remains an interesting topic.
Second, \ours does not guarantee that the integrated traffic rules are actually informative of the behavior of the agents. Measuring the alignment between informative priors and observations could help practitioners in selecting suitable task setups.
Third, our rule-filtering approach for generating synthetic training labels is based on static traffic rules. Natural language rules that require temporal evaluations are considerably more difficult to formalize using the RAG-LLM approach.
Fourth, prior knowledge about traffic rules must map to observable elements in the model input, such as road geometry and traffic signs. Missing relevant inputs (e.g., occluded traffic signs, unencoded lane types) prevent the model from learning anything informative from the synthetic training labels. This also applies, for example, to rasterized image inputs, where distinguishing finer details such as solid lines vs dashed lines might be challenging due to inherent limitations in the pixel resolution.
Despite these limitations, the modular design of \ours reveals an exciting path for future prior knowledge integration, as it can be scaled with the development of both rule formalization and prediction models. 

{\bf Acknowledgment.} 
The research leading to these results is funded by the German Federal Ministry for Economic Affairs and Energy within the project “NXT GEN AI METHODS – Generative Methoden für Perzeption, Prädiktion und Planung". We also thank Yue Yao for listening to our ideas and providing valuable feedback from a prediction perspective.
%\section{Acknowledgment}  
%\label{sec:ack} 
%This work is funded by the German Federal Ministry for Economic Affairs and Climate Action within the project ``nxtAIM”. We also thank Yue Yao for listening to our ideas and providing valuable feedback from a prediction perspective.

\bibliographystyle{plainnat}
\bibliography{references}
\newpage
\begin{appendices}

\section{Covariance Matrix $\kappa(h,h)$ and Estimation of Covariance in \ours}
\label{appendix:appendix_covariance_derivation}

\subsection{Random Fourier Features (RFF) Approximation}

Computing the exact GP covariance matrix $K(h,h)$ scales cubically with the number of data points $N$, making it computationally infeasible for large datasets. To mitigate this, we employ Random Fourier Features (RFF) to approximate the kernel function, thereby reducing the computational complexity.

\subsubsection{RFF for RBF Kernel}

The RBF kernel is defined as:

\begin{equation*}
    \kappa(h_i, h_j) = \exp\left(-\frac{\|h_i - h_j\|^2}{2\sigma^2}\right)
\end{equation*}

According to Bochner's theorem~\cite{rahimi2007bochner}, any shift-invariant kernel can be expressed as the Fourier transform of a non-negative measure. For the RBF kernel, the corresponding spectral distribution is Gaussian. Thus, the RFF approximation uses samples from this spectral distribution to construct random features.

\subsubsection{Construction of RFF}

Let $W \in \mathbb{R}^{d \times m}$ be a matrix whose entries are drawn from $\mathcal{N}(0, \frac{1}{\sigma^2})$, and $b \in [0, 2\pi]^m$ be a vector of biases uniformly drawn from $[0, 2\pi]$. The RFF approximation maps each input $h_i$ to a random feature vector $\phi(h_i) \in \mathbb{R}^m$ as follows:

\begin{equation*}
    \phi(h_i) = \sqrt{\frac{2}{m}} \cos(W h_i + b)
\end{equation*}

\subsubsection{Kernel Matrix Approximation}

Using RFF, the RBF kernel can be approximated by the inner product of the random features:

\begin{equation*}
    \kappa(h,h) \approx \Phi \Phi^\top
\end{equation*}

where $\Phi \in \mathbb{R}^{N \times m}$ is the matrix of random feature mappings for all inputs, i.e.,

\begin{equation*}
    \Phi = \begin{bmatrix}
    \phi(h_1)^\top \\
    \phi(h_2)^\top \\
    \vdots \\
    \phi(h_N)^\top
    \end{bmatrix}
\end{equation*}

Substituting the expression for $\phi(h_i)$, we get:

\begin{equation*}
\begin{split}
    \kappa(h,h) &\approx \Phi \Phi^\top \\
    &= \left(\sqrt{\frac{2}{m}} \cos(W H^\top + b \mathbf{1}^\top)\right) \\
    &\quad \times \left(\sqrt{\frac{2}{m}} \cos(W H^\top + b \mathbf{1}^\top)\right)^\top
\end{split}
\end{equation*}

where $H \in \mathbb{R}^{m \times N}$ is the matrix stacking all input embeddings $h_i$ as columns, and $\mathbf{1} \in \mathbb{R}^{1 \times N}$ is a vector of ones.

\subsection{Covariance Matrix Computation}

The covariance matrix computation involves the following steps:
\begin{enumerate}
    \item \textbf{Random Fourier Feature Mapping:}

    For each input embedding $h_i$, compute the random feature vector:

    \begin{equation*}
        \phi(h_i) = \sqrt{\frac{2}{m}} \cos(W h_i + b)
    \end{equation*}

 \item \textbf{Kernel Matrix Approximation:}

    Approximate the kernel matrix using the inner product of RFF:

    \begin{equation*}
        \kappa(h,h) \approx \Phi \Phi^\top
    \end{equation*}

    where $\Phi = [\phi(h_1), \phi(h_2), \dots, \phi(h_N)]^\top$.

\item \textbf{GP Posterior Covariance Estimation:}

    Using the Laplace approximation, estimate the posterior covariance matrix:

    \begin{equation*}
        \Sigma^{-1}_{g_c} \approx \mathbb{I} + \sum_{i=1}^N p_{i,c} (1 - p_{i,c}) \phi(h_i) \phi(h_i)^\top
    \end{equation*}

 \end{enumerate}

\section{Additional Results and Discussion}
\label{appendix:result_additional}

In our model's posterior estimation, the Gaussian Process (GP) posterior temperature hyperparameter regulates the trade-off between traffic rule-based priors and observed data. To understand its impact on trajectory prediction, we systematically varied the temperature from 5 to 45 and evaluated its influence on key performance metrics, including Negative Log-Likelihood (NLL), Average Displacement Error (ADE$_1$, ADE$_5$), Final Displacement Error (FDE$_1$), and Expected Calibration Error (ECE).

\begin{figure}[!htp]
\centering
    \includegraphics[width=\linewidth]{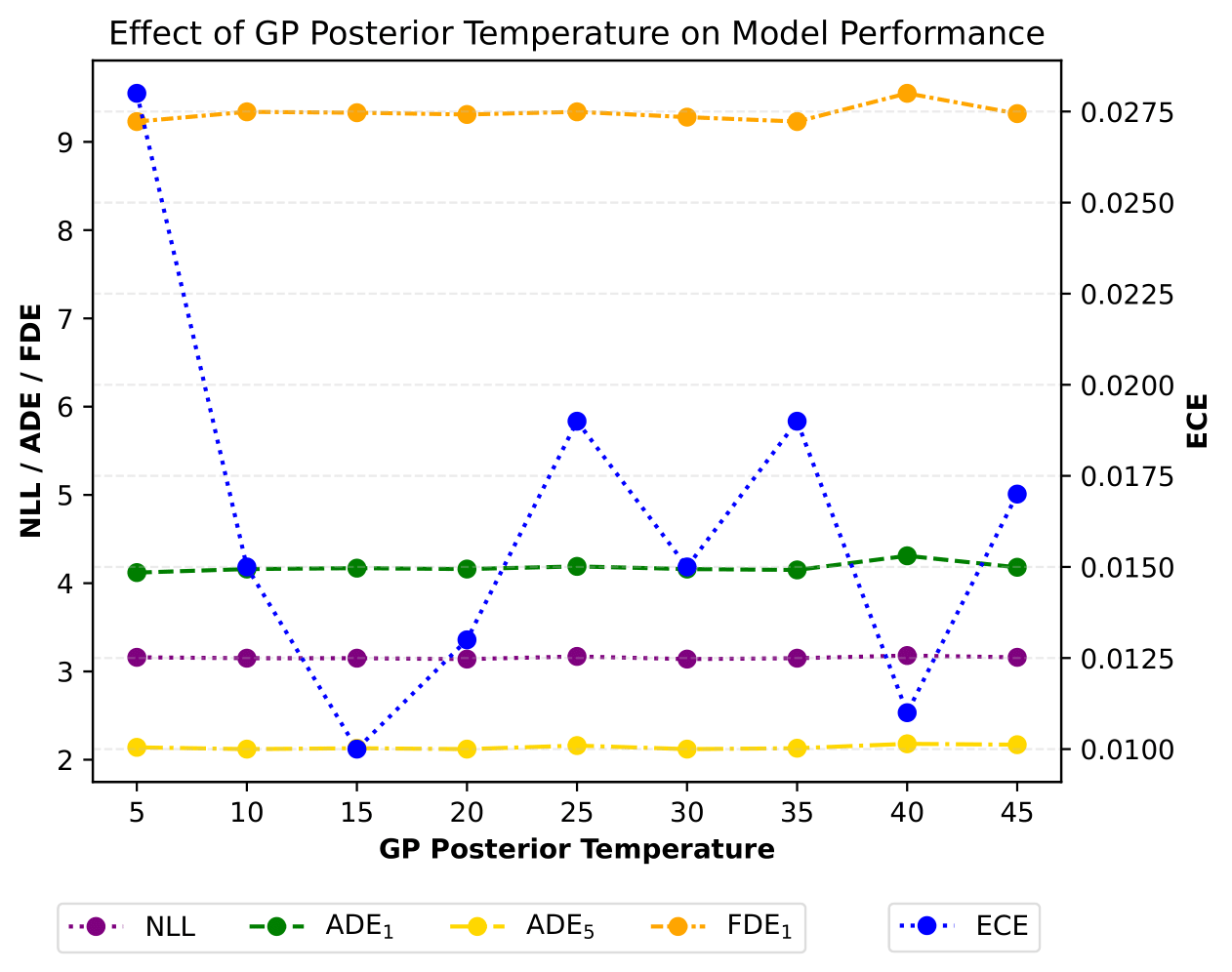}
    \caption{Impact of GP posterior temperature on trajectory prediction metrics. The x-axis shows temperature values (5-45), with the left y-axis displaying ADE$_1$, ADE$_5$, FDE$_1$, and NLL metrics, and the right y-axis showing ECE. Each metric is uniquely color-coded for visual distinction.}
    \label{fig:ablation:gp_temp}
\end{figure}

As illustrated in Fig.~\ref{fig:ablation:gp_temp}, our analysis shows that while NLL and ADE$_5$ remain relatively stable across temperature variations, model calibration fluctuates due to the inherently low baseline ECE values. Notably, temperatures outside the range of 15–30 degrade the accuracy of ADE$_1$ and ADE$_5$. We identify 15–30 as the optimal temperature range, striking a balance between predictive accuracy and uncertainty calibration.

This study highlights the significant role of the GP posterior temperature in governing \ours's ability to integrate prior knowledge with data-driven learning. Practitioners can optimize model performance for specific applications when tuned alongside other hyperparameters, such as the GP $\mathcal{L}_2$ regularizer. Additionally, the nature of the traffic rules incorporated into the prior plays a crucial role, as different rules impact prediction accuracy and calibration in distinct ways.

\section{LLM enabled Traffic Rules to Python code Generation and Limitations}
\label{appendix:llm_and_prompt}

To streamline the creation of traffic rule validation functions for trajectory prediction, we utilized a retrieval-augmented generation (RAG)~\cite{lewis2020rag} approach with a large language model (LLM) and prompting as described in~\citet{manas_tr2mtl,manas2024cot}. Specifically, the nuScenes map API\footnote{\href{https://github.com/nutonomy/nuscenes-devkit/tree/master/python-sdk/nuscenes/map_expansion}{https://github.com/nutonomy/nuscenes-devkit/tree/master/python-sdk/nuscenes/map\_expansion}} was employed as a contextual knowledge source. The map API provided access to high-definition (HD) maps, including information on pedestrian crossings, road boundaries, and traffic signs, which served as the basis for generating Python functions tailored to specific traffic rules. By incorporating this map-based context into the LLM prompts, the model generated executable Python code to enforce various constraints in the prior model labeling process.

For instance, when prompted to create a function for traffic rule ``\texttt{The vehicle trajectory should not cross pedestrian crossings in the presence of pedestrians}" the LLM generated the Python function as shown in Listing 1, which can be directly used in our prior model code base.

Our detailed Prompt file and setup for this is provided in attached code for SHIFT, which includes prompt template and single shot example for this specific task.

\begin{lstlisting}[language=Python, caption={LLM generated Function to check if trajectory crosses active pedestrian crossings}]
from shapely.geometry import LineString

def is_trajectory_crossing_pedestrian_crossings(trajectory, translation, rotation, map_name, nusc_map, active_crossings=None):
    """
    Check if the trajectory crosses any active pedestrian crossings.

    :param trajectory: List of points representing the trajectory in local coordinates.
    :param translation: Translation vector to convert local coordinates to global coordinates.
    :param rotation: Rotation angle to convert local coordinates to global coordinates.
    :param map_name: Name of the map being used (e.g., 'boston-seaport').
    :param nusc_map: NuScenesMap instance for accessing map data.
    :return: True if the trajectory crosses any active pedestrian crossings, False otherwise.
    """
    # Convert trajectory to global coordinates
    trajectory_global = convert_local_coords_to_global(
        trajectory,
        translation,
        rotation,
    )
    trajectory_line = LineString(trajectory_global)

    # Retrieve pedestrian crossings for the map
    if map_name not in nusc_map.ped_crossing:
        ped_crossing_records = nusc_map.ped_crossing
        ped_crossing_polygons = [
            nusc_map.explorer.extract_polygon(record['polygon_token'])
            for record in ped_crossing_records
        ]
    else:
        ped_crossing_polygons = nusc_map.ped_crossing[map_name]


    if active_crossings:
        ped_crossing_polygons = [
            nusc_map.explorer.extract_polygon(record['polygon_token'])
            for record in ped_crossing_records
            if record['token'] in active_crossings
        ]

    # Check for intersection with each pedestrian crossing
    for crossing_polygon in ped_crossing_polygons:
        if trajectory_line.crosses(crossing_polygon):
            return True  # The trajectory crosses an active pedestrian crossing

    return False  # No crossing 
\end{lstlisting}

This function integrates HD map data to determine whether a trajectory crosses active pedestrian crossings, showcasing the capability of the RAG-based LLM to generate high-quality, context-specific Python code. Similarly, other traffic rules defined in Sec.~\ref{sec:rules} were generated using this approach. While the RAG-based LLM provides substantial benefits in automating code generation, challenges such as hallucinations and incorrect or incomplete context remain prevalent in the process. 
RAG mitigates these issues to some extent by grounding the generated functions in real-world map sensor data provided by the nuScenes API, enhancing the accuracy and alignment of the outputs with the specified traffic rules. This allowed us to efficiently create prior model labels adhering to a wide range of traffic rules, including both global constraints (e.g., road boundaries) and local constraints (e.g., pedestrian priority). 

\textbf{Limitations and Mitigation of LLM-based generation.} we operate as a \textit{human-in-the-loop} workflow, requiring manual verification and refinement of the LLM-generated outputs. Limitations in the dataset also pose constraints. For instance, the dataset does not include information on emergency vehicles, preventing the generator from producing rules for such scenarios. Furthermore, failures occasionally arise due to the nuanced interpretation of natural language prompts. For example, the generated code used `shapely.crosses()' or `shapely.touches()' instead of `shapely.intersects()' based on the wording of the prompt, which can impact the accuracy of rule-compliant trajectory labels in the prior model. While these issues are not highly detrimental in our soft-prior setup, they underscore the need for careful review to ensure consistency and reliability in the generated outputs. Future work can explore solutions incorporating user-defined constraints and keyword selection or leveraging fine-tuned LLMs with low-rank adaptation. A human-in-the-loop system would significantly streamline the process, making it faster and more efficient by enabling users to verify rules rather than manually creating rule functions from scratch. This approach reduces the complexity of integrating map APIs and codebases for function generation, enhancing usability and scalability.

\section{Implementation Details}
\label{appendix:Implementation Details}
This section provides key implementation details of our model. For complete configurations and code, refer to our repository.
\subsection*{1. Uncertainty Estimation}
To balance epistemic and aleatoric uncertainty, we assign weighting factors of \( w_{\text{sngp}} = 0.1 \) and \( w_{\text{het}} = 0.2 \). The Gaussian Process (GP) kernel is approximated using 1,024 inducing points in Random Fourier Features (RFF), ensuring computational efficiency. Spectral normalization with a bound of 2.65 is applied to stabilize training and prevent gradient-related issues.

\subsection*{Temperature Parameters}

We introduce temperature parameters to control the influence of prior knowledge and regularization:
\begin{itemize}
    \item Temperature for GP posterior: 35, regulating influence of traffic rule prior in the GP layer.
    \item Temperature for feature extractor: 2.8, determining the effective dataset size for $L_2$ regularization.
\end{itemize}
These parameters ensure a balanced integration of prior knowledge and empirical observations.

\subsection*{2. Learning Rate Schedule}

The learning rate is adjusted based on the training data availability:
\begin{itemize}
    \item 100\% training data: \( 0.02 \).
    \item 50\% training data: \( 0.0005 \).
    \item 10\% training data: \( 0.0001 \).
\end{itemize}
A higher learning rate is used for larger datasets to accelerate convergence, while smaller datasets require lower rates to prevent overfitting.

\subsection*{3. Regularization and Early Stopping}

To improve generalization, we apply:
\begin{itemize}
    \item Early stopping with a patience of 15 epochs.
    \item $L_2$ regularization with a weight of 0.625 on the extractor output.
\end{itemize}

\subsection*{4. Parameter Search and Hyperparameter Tuning}
We utilize Bayesian optimization with Ray Tune\footnote{\href{https://docs.ray.io/en/latest/tune/index.html}{https://docs.ray.io/en/latest/tune/index.html}} to systematically search for optimal hyperparameters, including the learning rate, spectral norm bound, and feature extractor temperature.

\section{Additional Qualitative Results}
\label{appendix:qualitative_results}

\begin{figure}
    \centering
    \includegraphics[width=1\linewidth]{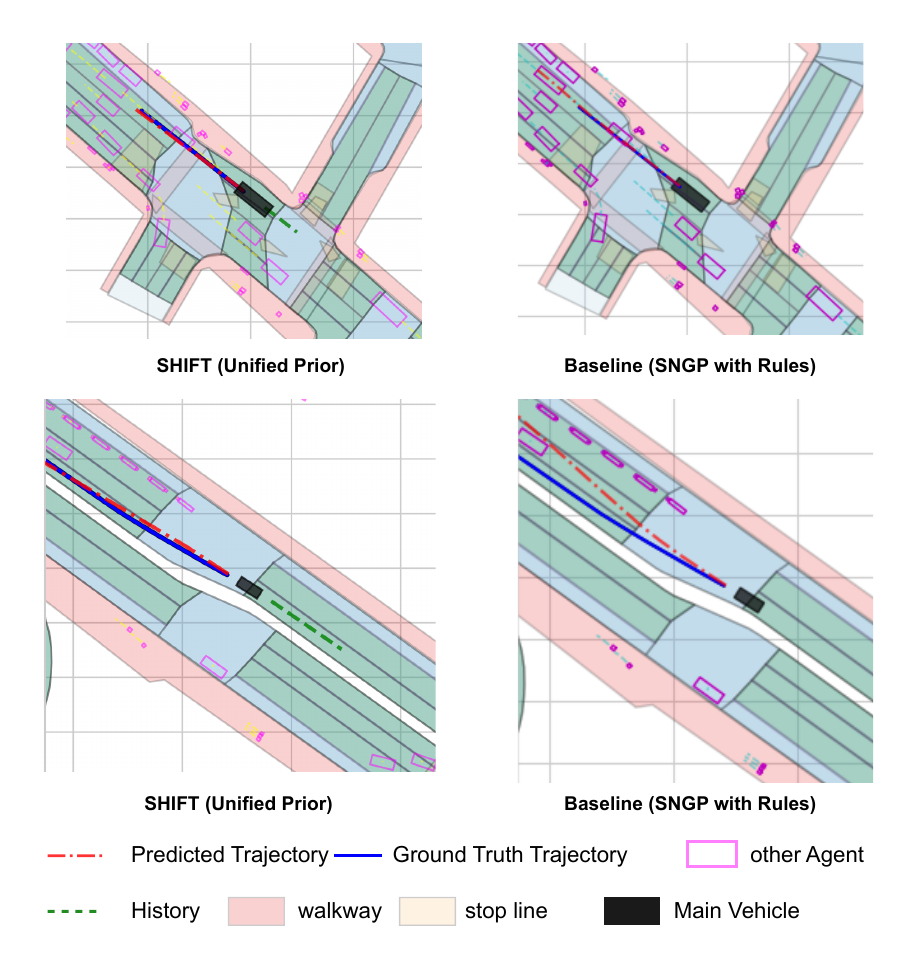}
    \caption{\ours demonstrates better prediction accuracy, closely following the ground truth trajectory in intersection crossing and lane-changing scenarios. In contrast, the baseline model predicts a trajectory with higher speed while crossing the intersection and exhibits lane-change behavior that is closer to another agent, making it potentially unsafe.}
    \label{fig:appendix_hetsngp_better}
\end{figure}

\begin{figure}
    \centering
    \includegraphics[width=1\linewidth]{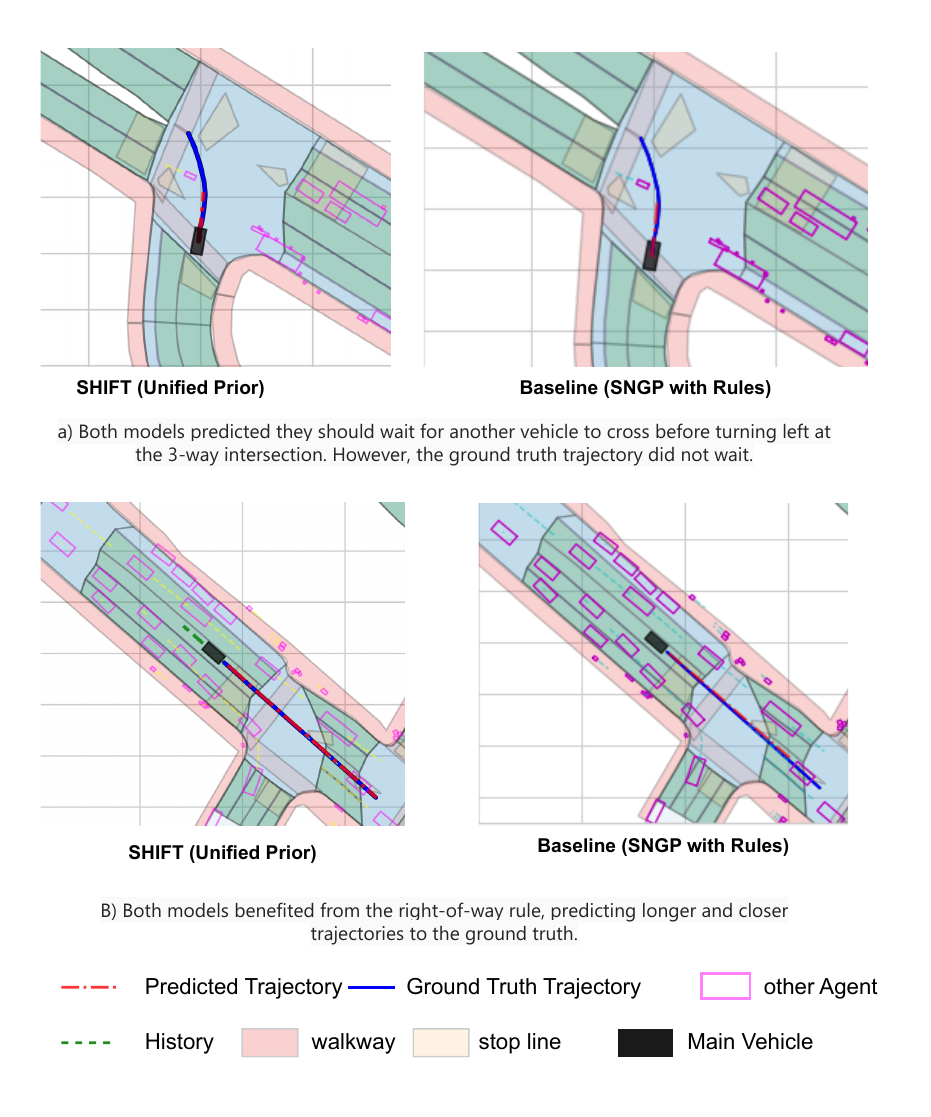}
    \caption{Both the baseline and \ours models perform similarly in these scenarios. In the top image, both models struggle with the decision to either wait for the approaching agent—who has priority—or proceed through the intersection, leading to slightly faster intersection crossing. In the bottom image, both models benefit from right-of-way rules, predicting longer trajectories that are closer to the ground truth. }
    \label{fig:appendix_both_better}
\end{figure}

In this section, we present additional qualitative results, highlighting not only the performance of \ours but also scenarios where both SNGP (with Rules) and \ours perform comparably well, even in complex situations, as well as instances where both methods encounter similar challenges.  

\textbf{Cases Where \ours Outperforms the Baseline.}
Figure \ref{fig:appendix_hetsngp_better} illustrates a scenario where \ours demonstrates superior prediction performance. In both intersection crossing and lane-changing scenarios, \ours closely follows the ground truth trajectory, whereas the baseline (SNGP with Rules) predicts a trajectory with higher speed while crossing the intersection. Additionally, during the lane-change maneuver, the baseline's prediction is closer to another agent, which could lead to a potentially unsafe interaction. These results suggest that \ours better captures motion dynamics and interactions with surrounding agents, resulting in more reliable predictions.  

\textbf{Cases Where Both Models Perform Similarly.}  
Figure \ref{fig:appendix_both_better} presents scenarios where both \ours and the baseline perform on par. In the top image, both models exhibit similar behavior when predicting a left turn at a three-way intersection. While the ground truth trajectory does not wait, both models opt to wait for another agent with priority before proceeding. This indicates that both models incorporate traffic rules effectively but may be overly cautious in certain cases. In the bottom image, both models benefit from right-of-way rules, predicting longer and more accurate trajectories that closely match the ground truth. These results highlight that in structured environments with clear right-of-way constraints, both models can generate reasonable predictions.  

Overall, these additional qualitative analyses demonstrate the strengths of \ours in improving trajectory prediction while also acknowledging cases where both models exhibit similar behavior and challenges.

\section{Dataset Statistics and Details}
\label{appendix:dataset}
The \textbf{nuScenes dataset} is a large-scale autonomous driving dataset designed for tasks such as 3D object detection, tracking, and trajectory prediction. It provides rich annotations and sensor data, making it suitable for trajectory prediction tasks. Below are the key statistics and details relevant to trajectory prediction:

\subsection*{1. General Dataset Overview}
\begin{itemize}
    \item \textbf{Total Scenes}: 1,000 scenes (each 20 seconds long).
    \item \textbf{Total Frames}: $\sim$40,000 keyframes (annotated at 2 Hz).
    \item \textbf{Sensor Data}: Consists of camera, RADAR and LiDAR sensors.
    \item \textbf{Geographic Diversity}: Collected in \textbf{Boston} and \textbf{Singapore}, covering diverse urban driving scenarios.
\end{itemize}

The dataset is divided into training, validation, and test splits, with a geographic distribution as follows:

\begin{itemize}
    \item \textbf{Total Data:}
    \begin{itemize}
        \item Train: 32,186 samples 
        \item Train-Val: 8,560 samples 
        \item Validation: 9,041 samples 
    \end{itemize}
    
    \item \textbf{Boston Subset:} 
    \begin{itemize}
        \item Train: 19,629 samples (60.99\% of total train sample)
        \item Train-Val: 5,855 samples (68.40\% of total train-val sample)
        \item Validation: 5,138 samples (56.84\% of validation sample)
    \end{itemize}
    
    \item \textbf{Singapore Subset:}
    \begin{itemize}
        \item Train: 12,557 samples (39.01\% of total train sample)
        \item Train-Val: 2,705 samples (31.60\% of total train-val sample)
        \item Validation: 3,903 samples (43.16\% of validation sample))
    \end{itemize}
\end{itemize}

The dataset is geographically diverse, with \textbf{Boston} representing North American driving conditions and \textbf{Singapore} representing Asian driving conditions. This diversity ensures that models trained on nuScenes generalize well to different regions.

\textbf{Annotations for Trajectory Prediction}
\begin{itemize}
    \item \textbf{Annotated Objects}: 1.4 million 3D bounding boxes across 23 object classes.
    \item \textbf{Relevant Classes for Trajectory Prediction}:
    \begin{itemize}
        \item Vehicles (car, truck, bus, trailer, etc.).
        \item Vulnerable road users such as pedestrians.
    \end{itemize}
    \item \textbf{Trajectory Annotations}:
    \begin{itemize}
        \item Each object has a 3D bounding box annotated at 2 Hz.
        \item Historical trajectories are available for each object (up to 2 seconds of past data).
        \item Future trajectories can be extrapolated for prediction tasks.
    \end{itemize}
\end{itemize}

\textbf{Trajectory Prediction-Specific Statistics.}
\begin{itemize}
    \item \textbf{Trajectory Length}: Future trajectory up to 6 seconds (12 frames at 2 Hz) for evaluation.

    \item \textbf{Interaction Scenarios}:
    \begin{itemize}
        \item Intersections, roundabouts, and lane changes are common, providing diverse interaction scenarios for trajectory prediction.
    \end{itemize}
\end{itemize}

\textbf{Challenges for Trajectory Prediction:}
The dataset presents challenges such as complex multi-agent interactions (e.g., vehicles and pedestrians), diverse driving scenarios (urban, highway, residential), and partial observations due to occlusions and limited sensor range, making trajectory prediction inherently difficult.
\end{appendices}

\end{document}